\pdfoutput=1

\documentclass[11pt]{article}

\usepackage{authblk}
\usepackage[]{acl}

\usepackage{times}
\usepackage{latexsym}

\usepackage[T1]{fontenc}

\usepackage[utf8]{inputenc}

\usepackage{microtype}

\usepackage{algorithm}
\usepackage{algpseudocode}
\usepackage{amsmath}

\usepackage{amssymb}
\usepackage{bm}
\usepackage{float}
\usepackage{booktabs}
\usepackage{adjustbox}
\usepackage{multirow}
\usepackage{makecell}
\usepackage{tabularx}
\usepackage[inline]{enumitem}
\usepackage{url}
\usepackage{graphicx}
\usepackage{subcaption}
\usepackage{colortbl}
\usepackage{courier}
\usepackage{mathtools}

\usepackage{tikz}
\usetikzlibrary{positioning, fit, mindmap, trees, calc,tikzmark,shapes}
\usetikzlibrary{shapes.arrows, fadings, automata,tikzmark,decorations.pathreplacing,patterns}
\usetikzlibrary{shapes.multipart}
\usetikzlibrary{arrows.meta}
\usetikzlibrary{intersections}
\usetikzlibrary{fit}
\usetikzlibrary{matrix}

\usepackage{pgfplots}
\usetikzlibrary{pgfplots.groupplots}
\usepackage{pgfplots,pgfplotstable}
\usetikzlibrary{patterns}

\usepackage{sistyle}
\SIthousandsep{,}

\setlist[itemize]{noitemsep, topsep=0pt}
\setlist[enumerate]{noitemsep, topsep=0pt}

\title{On Length Divergence Bias in Textual Matching Models}

\author[1]{Lan Jiang}
\author[2]{Tianshu Lyu}
\author[3]{Yankai Lin}
\author[2]{Meng Chong}
\author[2]{Xiaoyong Lyu}
\author[2]{Dawei Yin}
\affil[1]{Department of Automation, Tsinghua University}
\affil[2]{Baidu Inc., Beijing, China}
\affil[3]{Pattern Recognition Center, WeChat AI, Tencent Inc., China}
\affil[ ]{\texttt {jiangl20@mails.thu.edu yankailin@tencent.com}}
\affil[ ]{\texttt {\{lyutianshu, mengchong01, lvxiaoyong\}@baidu.com yindawei@acm.org}}

\begin{document}
\maketitle
\begin{abstract}
Despite the remarkable success deep models have achieved in Textual Matching (TM) tasks, it still remains unclear whether they truly understand language or measure the semantic similarity of texts by exploiting statistical bias in datasets.
In this work, we provide a new perspective to study this issue --- via the length divergence bias.
We find the length divergence heuristic widely exists in prevalent TM datasets, providing direct cues for prediction. 
To determine whether TM models have adopted such heuristic, we introduce an adversarial evaluation scheme which invalidates the heuristic.
In this adversarial setting, all TM models perform worse, indicating they have indeed adopted this heuristic.
Through a well-designed probing experiment, we empirically validate that the bias of TM models can be attributed in part to extracting the text length information during training.
To alleviate the length divergence bias, we propose an adversarial training method. 
The results demonstrate we successfully improve the robustness and generalization ability of models at the same time.
\end{abstract}

\begin{figure*}[t]
\centering
\resizebox{\linewidth}{!}{
\begin{tikzpicture}
\pgfplotsset{
    width=0.33\linewidth,
    height=0.23\linewidth,
    compat=1.17,
}

  \begin{groupplot}[
    group style={
        columns=4,
        group name=plots,
        x descriptions at=edge bottom,
        y descriptions at=edge left,
        horizontal sep=0.5cm,
    },
    ylabel={\# Examples},
    ybar,
    ymin=200,
    /pgf/bar width=6.5pt,
    symbolic x coords={CAT1, CAT2, CAT3, CAT4},
    xtick=data,
    label style={font=\footnotesize},
    tick label style={font=\footnotesize},
    legend columns=3]

    \nextgroupplot[title=QQP, legend to name=grouplegend]
    \addplot[area legend, pattern=dots]
	coordinates
	{
	 (CAT1,37055) (CAT2,36115) (CAT3,38394) (CAT4,22814)
	};\label{pos}
	\addplot[area legend, fill = {rgb:black,1;white,2}]
	coordinates
	{
	 (CAT1,51886) (CAT2,43223) (CAT3,62115) (CAT4,72247) 
	};\label{neg}

    \nextgroupplot[title=Twitter-URL, scaled y ticks=false]
    \addplot[area legend, pattern=dots]  
	coordinates
	{
	 (CAT1,3036) (CAT2,2551) (CAT3,2712) (CAT4,2868)
	};\label{pos}
	\addplot[area legend, fill = {rgb:black,1;white,2}]   
	coordinates
	{
	 (CAT1,6549) (CAT2,6019) (CAT3,8262) (CAT4,10203) 
	};\label{neg}

    \nextgroupplot[title=TrecQA, scaled y ticks=false]
    \addplot[area legend, pattern=dots]
	coordinates
	{
	 (CAT1,2146) (CAT2,2295) (CAT3,1315) (CAT4,647)
	};\label{pos}
	\addplot[area legend, fill = {rgb:black,1;white,2}]
	coordinates
	{
	 (CAT1,13621) (CAT2,15986) (CAT3,11792) (CAT4,5615) 
	};\label{neg}

	\nextgroupplot[title=Microblog, scaled y ticks=false]
    \addplot[area legend, pattern=dots]
	coordinates
	{
	 (CAT1,1582) (CAT2,857) (CAT3,705) (CAT4,841)
	};\label{pos}
	\addplot[area legend, fill = {rgb:black,1;white,2}]
	coordinates
	{
	 (CAT1,7823) (CAT2,8440) (CAT3,9786) (CAT4,9344) 
	};\label{neg}
  \end{groupplot}

  \begin{groupplot}[
  	group style= {
        columns=4,
        group name=plots,
        y descriptions at=edge right,
        horizontal sep=0.5cm,
    },
    ymin=0,
    symbolic x coords={CAT1, CAT2, CAT3, CAT4},
    axis x line=none,
    ylabel={PosRatio (\%)},
    label style={font=\footnotesize},
    tick label style={font=\footnotesize},
    legend columns=3,
    ]

    \nextgroupplot[scaled y ticks=false, legend to name=grouplegend]
    \addlegendimage{/pgfplots/refstyle=pos}\addlegendentry{Positive}
    \addlegendimage{/pgfplots/refstyle=neg}\addlegendentry{Negative}
    \addplot[mark=x]
	coordinates                         
	{                                   
	 (CAT1,42) (CAT2,46) (CAT3,38) (CAT4,24)
	};\addlegendentry{PosRatio}

    \nextgroupplot[scaled y ticks=false]
   	\addplot[mark=x]              
	coordinates                         
	{                                   
	 (CAT1,32) (CAT2,30) (CAT3,25) (CAT4,22)
	};\label{pos ratio}

	\nextgroupplot[scaled y ticks=false]
    \addplot[mark=x]              
	coordinates                         
	{                                   
	 (CAT1,16) (CAT2,14) (CAT3,11) (CAT4,12)
	};\label{pos ratio}
    
    \nextgroupplot
    \addplot[mark=x]              
	coordinates                         
	{                                   
	 (CAT1,20) (CAT2,10) (CAT3,7.2) (CAT4,9)
	};\label{pos ratio}
  \end{groupplot}

  \node at (plots c3r1.south) [anchor=south, xshift=-0.12\linewidth, yshift=-1.5cm] {
  \ref{grouplegend}
  };
\end{tikzpicture}
}
\caption{Length divergence distribution by labels across datasets. Bars represent the number of examples, corresponding to the left axis; polylines represent the ratio of positive examples, corresponding to the right axis.}
\label{fig:length-cues}
\end{figure*}
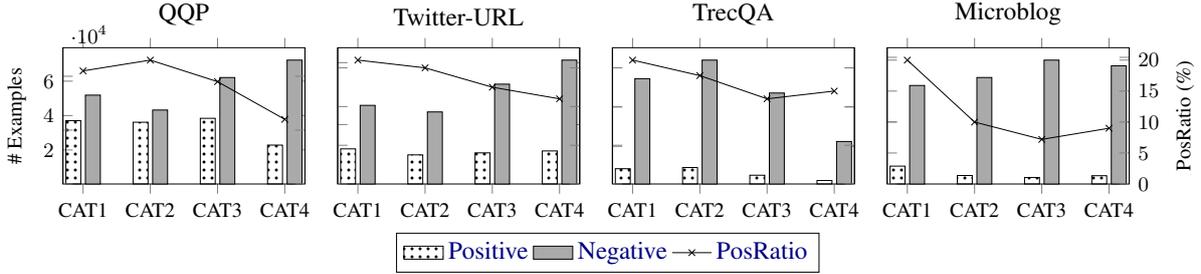
\section{Introduction}\label{section:intro}
Textual matching is a crucial component in various NLP applications, such as information retrieval \cite{10.1561/1500000035}, question answering \cite{shen-lapata-2007-using} and duplicate detection \cite{10.1145/956750.956759}. 
Given a pair of texts, the goal is to determine the semantic similarity between them.
A lot of deep models \cite{chen_enhanced_2017,wang_bilateral_2017,pang_text_2016,Guo:2019:MLP:3331184.3331403,10.5555/3016100.3016298} have achieved excellent performance on various TM benchmarks. 

However, recent work has found that current models are prone to adopting shallow heuristics in the datasets, rather than learning the underlying linguistics that they are intended to capture.
This issue has been documented across tasks in natural language understanding.
In natural language arguments, for example, \citet{niven_probing_2019} showed that model performance is inflated by spurious statistical cues.
Similar heuristics arise in natural language inference \cite{mccoy-etal-2019-right,naik-etal-2018-stress} and reading comprehension \cite{jia_adversarial_2017}.

\begin{table}[t]
    \centering
    \resizebox{\linewidth}{!}{
    \begin{tabular}{p{7.6cm}}
        \toprule
        $T_1$ - Microsoft acquires Maluuba, a startup focused on general artificial intelligence. (\textbf{10})\\
        $T_2$ - Microsoft has acquired Canadian startupMaluuba, a company founded by University of Waterloo grads Kaheer Suleman and Sam Pasupalak that also participated in. (\textbf{22})\\
        Label: \textit{\textcolor{red}{Paraphrase}}\hspace{0.5cm}Output: \textit{\textcolor{blue}{Non-paraphrase}}\\
        \midrule
        $T_1$ - Bill would cut off aid to countries that don't take back their illegal immigrant criminals. (\textbf{15})\\
        $T_2$ - Common Sense law faces massive opposition supposing that Aid would be cut off to countries who refuse their citizens. (\textbf{19})\\
        Label: \textit{\textcolor{blue}{Non-paraphrase}}\hspace{0.5cm}Output: \textit{\textcolor{red}{Paraphrase}}\\
    \bottomrule
    \end{tabular}}
    \caption{Examples for length divergence bias, originally from Twitter-URL. "Output" is the output label by ESIM trained on the original training set. Numbers in bold are the number of words each text consists of. Model is misled by the length divergence of two texts.}
    \label{tab:error-case}%
\end{table}

In this paper, we address this issue in the domain of textual matching. 
The focus of our work is on the length divergence bias --- models tend to classify examples with high length divergence as negative and vice versa.
Table \ref{tab:error-case} shows a single set of instances from Twitter-URL that demonstrates the length divergence bias.

We analyze current TM datasets and find that all of them follow specific length divergence distribution by labels. 
To determine whether TM models have employed this spurious pattern to facilitate their performance, we construct adversarial test sets which invalidate this heuristic and re-evaluate TM models. 
There is a performance drop on 14 out of total 16 combinations of models and tasks, suggesting their reliance on this heuristic. 

Despite demonstrating the existence of length divergence bias, the underlying reason has not been well explained. 
By conducting the \textit{SentLen} probing experiment \cite{conneau_what_2018}, we bridge this gap through revealing the text length information TM models have learned during training.

We finally explore a simple yet effective adversarial training method to correct the length divergence bias. 
The results show our approach not only reduces the bias but also improves the generalization ability of TM models.
To encourage the development of TM models that understand semantics more precisely, we will release our code.
\section{Datasets and Models}

We select four well-known NLP and IR datasets as follows: 
\textbf{Quora Question Pairs (QQP)} \citep{wang_glue_2018}, 
\textbf{Twitter-URL} \citep{lan_continuously_2017}, 
\textbf{TrecQA} \cite{wang-etal-2007-jeopardy}, and
\textbf{TREC Microblog 2013 (Microblog)} \cite{DBLP:conf/trec/LinE13}. 

We study four models for textual matching tasks: 
\textbf{MatchPyramid} \cite{pang_text_2016}, \textbf{BiMPM} \cite{wang_bilateral_2017}, \textbf{ESIM} \cite{chen_enhanced_2017} and \textbf{BERT} \cite{devlin_bert_2019}. 
The four models above are representative in terms of neural architectures.

The detailed explanation for each dataset and model can be found in Appendix \ref{app-1} and \ref{app-2}.


\section{Length Divergence Heuristic in Current Datasets}

In this section, we characterize existing datasets from the perspective of the length divergence between text pairs. 
We first formulate pairwise length divergence for NLP tasks and listwise length divergence for IR tasks, respectively. 

\textbf{Pairwise.} Given two texts $T_1$ and $T_2$, their relative length divergence is defined as:
\begin{eqnarray}
  \Delta_{\textrm{rel}}\mathcal{L}(T_1,T_2) \coloneqq \frac{|\mathcal{L}_{T_1} - \mathcal{L}_{T_2}|}{\min(\mathcal{L}_{T_1}, \mathcal{L}_{T_2})} \label{eq:rel_len},
\end{eqnarray}
where 
\begin{eqnarray}
  \mathcal{L}_{T} \coloneqq \# (\textrm{words in } T).
\end{eqnarray}

\textbf{Listwise.} In IR tasks, each example consists of a query $Q$ and a list of documents $\bm{D}$ associated with it. 
We define the listwise relative length divergence with respect to $Q$ as:
\begin{eqnarray}
\resizebox{.85\hsize}{!}{$
\Delta_{\textrm{rel}}\mathcal{L}(Q,\bm{D}) \coloneqq \frac{|\overline{\Delta_{\textrm{rel}}\mathcal{L}}(Q,\bm{D^+})-\overline{\Delta_{\textrm{rel}}\mathcal{L}}(Q,\bm{D^-})|}{\min(\overline{\Delta_{\textrm{rel}}\mathcal{L}}(Q,\bm{D^+}), \overline{\Delta_{\textrm{rel}}\mathcal{L}}(Q,\bm{D^-}))}
$}, \\
\resizebox{.85\hsize}{!}{$
\overline{\Delta_{\textrm{rel}}\mathcal{L}}(Q,\bm{D^{+/-}})=\frac{\sum_{d^{+/-}\in \bm{D^{+/-}}}\Delta_{\textrm{rel}}\mathcal{L}(Q,d^{+/-})}{|\bm{D^{+/-}}|}
$},
\end{eqnarray}
where $\bm{D^+}$ is the set of relevant documents while $\bm{D^-}$ C
For instances whose $\overline{\Delta_{\textrm{rel}}\mathcal{L}}(Q,\bm{D^{+/-}})$ does not exist or is equal to zero, we set $\Delta_{\textrm{rel}}\mathcal{L}(Q,\bm{D})$ to be a large number.

Based on the length divergence definition, we sort and split the training sets into quarters, namely \texttt{CAT1-4}, and examine length divergence distribution by labels for each dataset. 
Statistics are shown in Figure \ref{fig:length-cues}. 
We can see that all datasets suffer from the same problem: as the length divergence increases, the ratio of positive examples decreases. 
Overall, negative examples tend to have higher length divergence than positive ones, providing direct cues for label assignment.
\section{Length Divergence Bias in TM Models}\label{sec:exp}

In this section, we demonstrate that existing TM models indeed employ the length divergence heuristic in datasets.
Existing test sets are drawn from the same distribution as the training sets, which are overly lenient on models that rely on superficial cues. 
To provide a more robust assessment of TM models, we construct adversarial test sets by eliminating such heuristic. 

\subsection{Adversarial Test Sets}

\begin{table*}[thb]
\begin{tabular}{lccccc|ccccc}
\toprule
\multicolumn{1}{l|}{\multirow{2}{*}{}} & \multicolumn{5}{c|}{Original} & \multicolumn{5}{c}{Adversarial} \\
\multicolumn{1}{c|}{}           & CAT1  & CAT2  & CAT3   & CAT4 & ALL       & CAT1  & CAT2  & CAT3   & CAT4 & ALL  \\ \hline
\multicolumn{1}{l|}{\# Positive}  & 4,055 & 3,967 & 4,280  & 2,583 & 14,885 & 3,385 & 2,882 & 4,013  & 2,583 & 12,863  \\
\multicolumn{1}{l|}{\# Negative} & 5,781 & 4,923 & 6,853  & 7,988 & 25,545 & 5,781 & 4,923 & 6,853  & 4,410  & 21,967  \\
\multicolumn{1}{l|}{\# Total}    & 9,836 & 8,890 & 11,133 & 10,571 & 40,430 & 9,166 & 7,805 & 10,866 & 6,993 & 34,830  \\ \hline
\multicolumn{1}{l|}{PosRatio}    & 0.41  & 0.45  & 0.38   & 0.24 & 0.37   & 0.37   & 0.37  & 0.37  & 0.37   & 0.37  \\ 
\bottomrule
\end{tabular}
\caption{Statistics of the original and adversarial test set on QQP task. Each category in the original test set is down-sampled to align with the average PosRatio.}
  \label{tab:adv-test}
\end{table*}

For two NLP datasets, adversarial test sets are built by the following steps: First, examples are sorted and split into four categories according to their length divergence. 
Second, we down-sample each category to align with the average PosRatio of the whole test set, i.e., the adversarial datasets we construct are subsets of theoriginal ones. 
Table \ref{tab:adv-test} gives the details of the adversarial test set we build on QQP task, with a comparison of the original one.

The construction of IR datasets follows the same first step as NLP datasets. 
Considering random down-sampling may break the completion of query and its associated documents, in the second step, we discard the fourth category directly instead of down-sampling across all categories. 

\subsection{Re-evaluating TM Models}

To examine whether TM models exploit the length divergence heuristic of existing datasets, models trained on the original training sets are evaluated on the original and adversarial test sets, respectively. 
We provide further details to facilitate reproducibility in Appendix \ref{app-3}.


\textbf{Results.} The results are shown in Table \ref{tab:adv-eval}. Overall, almost all models have a performance drop on all datasets (14 out of total 16 combinations). It seems that MatchPyramid captures the richest length divergence cues, as its performance drops most dramatically. 
BiMPM and ESIM both perform worse on the adversarial test sets except for one task. 
Although BERT outperforms other models, it cannot maintain its performance under adversarial evaluation either.

\begin{table*}[!htb]
\resizebox{\linewidth}{!}{
\begin{tabular}{l|c|cc|cc|cc|cc}
\toprule
\multirow{2}{*}{Datasets} & \multirow{2}{*}{Metrics} & \multicolumn{2}{c|}{MatchPyramid} & \multicolumn{2}{c|}{BiMPM} & \multicolumn{2}{c|}{ESIM} & \multicolumn{2}{c}{BERT} \\
& & Original & Adversarial & Original & Adversarial & Original & Adversarial & Original & Adversarial \\ \hline
\multirow{2}{*}{QQP} & Acc & 70.18 (+6.29) & 68.66 (+7.31) & 81.52 (-0.08) & 80.91 (+0.15) & 82.15 (-0.50) & 81.38 (-0.15) & 83.51 (+1.34) & 82.57 (+1.67) \\ 
& BA & 66.00 (+8.41) & 64.60 (+9.44) &  79.43 (+0.63) & 78.97 (+0.81) & 80.62 (-1.23) & 80.01 (-0.96) & 84.46 (+0.77) & 83.65 (+1.04) \\ \hline
\multirow{2}{*}{Twitter-URL} & macro-F1 & 72.28 (+0.36) & 71.72 (+0.38) & 77.94 (+0.20) & 77.63 (+0.17) & 76.42 (+0.90) & 75.91 (+1.21) & 80.30 (+0.49) & 80.10 (+0.55)  \\ 
& micro-F1 & 84.23 (-0.26) & 83.99 (-0.25) & 85.50 (+0.12) & 85.33 (+0.08) & 86.58 (-0.46) & 86.33 (-0.32) & 85.26 (+0.50) & 85.12 (+0.54)  \\ \hline
\multirow{2}{*}{TrecQA} & MAP & 60.22 (+6.18) & 57.88 (+8.20) & 88.75 (+2.24) & 91.64 (+2.10) & 76.84 (+7.18) & 77.74 (+8.26) & 87.22 (+1.31) & 89.61 (+0.35) \\ 
& MRR & 48.42 (+3.31) & 47.27 (+5.95) & 67.27 (+0.07) & 67.56 (+0.13) & 63.74 (-0.46) & 60.99 (+3.06) & 67.76 (-0.59) & 65.75 (+0.12) \\ \hline
Microblog & MAP & 18.93 (+0.23) & 15.75 (+0.65) & 26.44 (+15.06) & 25.30 (+12.13) & 14.54 (+22.61) & 17.84 (+12.01) & 47.11 (+2.79) & 38.15 (+1.76)  \\ 
\bottomrule
\end{tabular}}
\caption{Performances of four TM models on the original and adversarial test sets for four tasks. Models are first trained on the original training sets. Performances of all systems drop on the adversarial test sets compared to on the original test sets. Models are then re-trained on the adversarial training sets. Numbers in parentheses indicate absolute gains from adversarial training.}
  \label{tab:adv-eval}\textbf{}
\end{table*}

\begin{figure}[htp]
\centering
\resizebox{\linewidth}{!}{
\begin{tikzpicture}[font=\footnotesize]
\pgfplotsset{
    scale only axis,
}
\begin{axis}[
    ymajorgrids=true,
    width=0.8\linewidth,
    height=0.3\linewidth,
    symbolic x coords={CAT1, CAT2, CAT3, CAT4},
    xtick=data,
    ylabel=Recall,
    label style={font=\footnotesize},
    tick label style={font=\footnotesize},
    legend style={font=\tiny,at={(0,0)},anchor=south west,legend columns=2},
    ]
\addplot[mark=x,draw=blue]
coordinates
{
 (CAT1,0.5601) (CAT2,0.6108) (CAT3,0.4658) (CAT4,0.3115)
};\label{MatchPyramid}
\addlegendentry{MatchPyramid}
\addplot[mark=square,draw=orange]
coordinates
{
 (CAT1,0.709) (CAT2,0.7326) (CAT3,0.7288) (CAT4,0.6882)
};\label{BiMPM TPR}
\addlegendentry{BiMPM}
\addplot[mark=o,draw=red]
coordinates
{
 (CAT1,0.7751) (CAT2,0.7689) (CAT3,0.7448) (CAT4,0.6936)
};\label{ESIM TPR}
\addlegendentry{ESIM}
\addplot[mark=triangle,draw=green]
coordinates
{
 (CAT1,0.9028) (CAT2,0.9012) (CAT3,0.875) (CAT4,0.824)
};\label{BERT TPR}
\addlegendentry{BERT}
\end{axis}
\end{tikzpicture}
}
\caption{Recall of all models on the QQP adversarial test set. \texttt{CAT1} is the category with the lowest length divergence. Recall decreases across four categories, indicating TM models tend to determine examples with high length divergence as negative and vice versa.}
\label{fig:recall}
\end{figure}
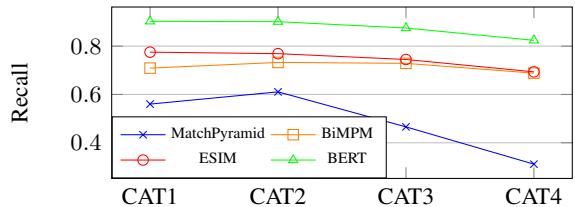

Moreover, we explore how the recall varies with the length divergence of examples. 
We report the recall for four length divergence categories of all models on QQP adversarial test set. 
Figure \ref{fig:recall} shows that the recall declines across four categories, which indicates that TM models are more inclined to determine examples with high length divergence as negative, and vice versa.

We address that the adversarial evaluation scheme, which invalidates the length divergence heuristic, provides a more robust assessment for TM models. 
The above results are well apt with our intuitions about the length divergence bias: models do exploit some superficial cues about length divergence, instead of truly understanding the meaning of texts despite their good performance. 

\subsection{Probing Experiment}

The adversarial evaluation has revealed the length divergence bias of TM models, but reason for this phenomenon is still unclear. 
In this section, we dig deeper into this problem. 

Despite the variations of the architectures of TM models, all of them need to extract representation of texts first. 
The linguistic properties TM models capture have a direct effect on the downstream tasks. 
To explore what kind of information TM models have learned, we introduce a probing experiment using representations produced by TM models to predict the length of texts. 
We conduct the \textit{SentLen} task in SentEval \cite{conneau_what_2018}, which is a \num{6}-way classification task performed by a simple MLP with Sigmoid activation. 
As MatchPyramid cannot produce representation of a single text, we do not include it in this probing experiment. 
BiMPM and ESIM model both employ a BiLSTM encoder. 
We select the maximum value over each dimension of the hidden units as text representations, and use untrained encoders with random weights as the baseline. 
As BERT uses the output of the first token (\texttt{[CLS]} token) to make classification, we report the classification result using \texttt{[CLS]} token representations. 
We use the pre-trained model without fine-tuning as baseline.

\begin{table}[tbp]
\resizebox{\linewidth}{!}{
\begin{tabular}{l|cccc}
\toprule
Models & QQP & Twitter-URL & TrecQA & Microblog \\ \hline
\hline \multicolumn{5}{l}{\textit{BiMPM}} \\ \hline
Untrained  & 56.18 & 54.47 & 56.23 & 57.03 \\ \hline
BiLSTM-max & 63.85 & 64.66 & 57.35 & 57.64 \\ \hline
\hline \multicolumn{5}{l}{\textit{ESIM}} \\ \hline
Untrained  & 58.28 & 56.99 & 55.81 & 56.89  \\ \hline
BiLSTM-max & 65.83 & 65.43 & 66.59 & 66.06  \\ \hline
\hline \multicolumn{5}{l}{\textit{BERT}} \\ \hline
Untrained  & \multicolumn{4}{c}{72.63}  \\ \hline
BERT-\texttt{[CLS]} & 70.66 & 72.10 & 81.33 & 75.08   \\
\bottomrule
\end{tabular}}
\caption{Probing task accuracies. Classifier takes text representation produced by TM models as input.}
\label{tab:probing-acc}
\end{table}

\textbf{Results.} From Table \ref{tab:probing-acc} we can see that BiLSTM encoder has a severe performance boost across all datasets after training, which indicates that BiMPM and ESIM extract rich text length information during training.
Compared to untrained BiLSTM encoder, BERT achieves a substantially better performance without fine-tuning. 
Interestingly, while BERT model suffers bias too, they are less pronounced, perhaps a benefit of the exposure to large corpus where the spurious patterns may not have held. 
It seems that pre-training enables BERT to extract relatively deeper linguistic properties and forget about superficial information.

The \textit{SentLen} probing experiment reveals that TM models learn rich text length information during training, indicating the intrinsic reason why TM models suffer from the length divergence bias.
\section{Length Divergence Bias Correction}

In this section, we propose to correct the length divergence bias. 
As the model modification method usually has a significant cost and is inefficient, we apply adversarial training with bias-free training data. 
Our method is much more practical, lower cost, and easier to be implemented and adopted. 
We first construct the adversarial training sets in the same way as the adversarial test sets. 
We next re-train each model on the adversarial training sets and report their performance on two test sets. 


\textbf{Results.} As presented in Table \ref{tab:adv-eval}, performances improve for almost all models across four datasets except for one combination (ESIM on QQP).
It is a little inspiring that adversarial training brings tremendous benefits to some models on IR tasks (MatchPyramid and ESIM on TrecQA dataset, BiMPM and ESIM on Microblog dataset). 
One possible explanation for this phenomenon is that, in IR tasks, the length divergence and class-imbalance are more severe than NLP tasks.
While alleviating the length divergence bias of TM models, our method also makes models achieve better performances on the original test sets.
Overall, the adversarial training not only successfully corrects the length divergence bias in TM models but also improves their generalization ability. 
\section{Conclusion}
The inspiring success of deep models is accounted for by employment spurious heuristics in datasets, instead of truly understanding the language.
In this work, we investigate the length divergence heuristic that textual matching models are prone to learn.
We characterize current TM datasets and find that examples with high length divergence tend to have negative labels and vice versa. 
To provide a more robust assessment, we construct adversarial test sets, on which models using this heuristic are guaranteed to fail.
Experiments show that almost all TM models perform worse on adversarial test sets, indicating they indeed exploit the length divergence cues. 
We then provide a deeper insight by conducting the \textit{SentLen} probing experiment. 
TM models are shown to learn rich text length information during training, which accounts for the length divergence bias.
Finally, we propose a simple yet effective adversarial training method to alleviate the length divergence bias in TM models.
It's a little inspiring that our approach improves models' robustness and generalization ability at the same time. 
Overall, our results indicate that, there is still substantial room towards TM models which understand language more precisely.

\bibliography{anthology}

\begin{thebibliography}{24}
\expandafter\ifx\csname natexlab\endcsname\relax\def\natexlab#1{#1}\fi

\bibitem[{Bilenko and Mooney(2003)}]{10.1145/956750.956759}
Mikhail Bilenko and Raymond~J. Mooney. 2003.
\newblock Adaptive duplicate detection using learnable string similarity
  measures.
\newblock In \emph{Proceedings of the Ninth ACM SIGKDD International Conference
  on Knowledge Discovery and Data Mining}, KDD '03, page 39–48, New York, NY,
  USA. Association for Computing Machinery.

\bibitem[{Chen et~al.(2017)Chen, Zhu, Ling, Wei, Jiang, and
  Inkpen}]{chen_enhanced_2017}
Qian Chen, Xiaodan Zhu, Zhen-Hua Ling, Si~Wei, Hui Jiang, and Diana Inkpen.
  2017.
\newblock Enhanced {LSTM} for {Natural} {Language} {Inference}.
\newblock In \emph{Proceedings of the 55th {Annual} {Meeting} of the
  {Association} for {Computational} {Linguistics} ({Volume} 1: {Long}
  {Papers})}, pages 1657--1668, Vancouver, Canada. Association for
  Computational Linguistics.

\bibitem[{Conneau et~al.(2018)Conneau, Kruszewski, Lample, Barrault, and
  Baroni}]{conneau_what_2018}
Alexis Conneau, German Kruszewski, Guillaume Lample, Loïc Barrault, and Marco
  Baroni. 2018.
\newblock What you can cram into a single \$\&!\#* vector: {Probing} sentence
  embeddings for linguistic properties.
\newblock In \emph{Proceedings of the 56th {Annual} {Meeting} of the
  {Association} for {Computational} {Linguistics} ({Volume} 1: {Long}
  {Papers})}, pages 2126--2136, Melbourne, Australia. Association for
  Computational Linguistics.

\bibitem[{Devlin et~al.(2019)Devlin, Chang, Lee, and
  Toutanova}]{devlin_bert_2019}
Jacob Devlin, Ming-Wei Chang, Kenton Lee, and Kristina Toutanova. 2019.
\newblock {BERT}: {Pre}-training of {Deep} {Bidirectional} {Transformers} for
  {Language} {Understanding}.
\newblock In \emph{Proceedings of the 2019 {Conference} of the {North}
  {American} {Chapter} of the {Association} for {Computational} {Linguistics}:
  {Human} {Language} {Technologies}, {Volume} 1 ({Long} and {Short} {Papers})},
  pages 4171--4186, Minneapolis, Minnesota. Association for Computational
  Linguistics.

\bibitem[{Guo et~al.(2019)Guo, Fan, Ji, and
  Cheng}]{Guo:2019:MLP:3331184.3331403}
Jiafeng Guo, Yixing Fan, Xiang Ji, and Xueqi Cheng. 2019.
\newblock Matchzoo: A learning, practicing, and developing system for neural
  text matching.
\newblock In \emph{Proceedings of the 42Nd International ACM SIGIR Conference
  on Research and Development in Information Retrieval}, SIGIR'19, pages
  1297--1300, New York, NY, USA. ACM.

\bibitem[{Jia and Liang(2017)}]{jia_adversarial_2017}
Robin Jia and Percy Liang. 2017.
\newblock Adversarial {Examples} for {Evaluating} {Reading} {Comprehension}
  {Systems}.
\newblock In \emph{Proceedings of the 2017 {Conference} on {Empirical}
  {Methods} in {Natural} {Language} {Processing}}, pages 2021--2031,
  Copenhagen, Denmark. Association for Computational Linguistics.

\bibitem[{Kingma and Ba(2015)}]{DBLP:journals/corr/KingmaB14}
Diederik~P. Kingma and Jimmy Ba. 2015.
\newblock Adam: {A} method for stochastic optimization.
\newblock In \emph{3rd International Conference on Learning Representations,
  {ICLR} 2015, San Diego, CA, USA, May 7-9, 2015, Conference Track
  Proceedings}.

\bibitem[{Lan et~al.(2017)Lan, Qiu, He, and Xu}]{lan_continuously_2017}
Wuwei Lan, Siyu Qiu, Hua He, and Wei Xu. 2017.
\newblock A {Continuously} {Growing} {Dataset} of {Sentential} {Paraphrases}.
\newblock In \emph{Proceedings of the 2017 {Conference} on {Empirical}
  {Methods} in {Natural} {Language} {Processing}}, pages 1224--1234,
  Copenhagen, Denmark. Association for Computational Linguistics.

\bibitem[{Li and Xu(2014)}]{10.1561/1500000035}
Hang Li and Jun Xu. 2014.
\newblock Semantic matching in search.
\newblock \emph{Found. Trends Inf. Retr.}, 7(5):343–469.

\bibitem[{Lin and Efron(2013)}]{DBLP:conf/trec/LinE13}
Jimmy~J. Lin and Miles Efron. 2013.
\newblock Overview of the {TREC-2013} microblog track.
\newblock In \emph{Proceedings of The Twenty-Second Text REtrieval Conference,
  {TREC} 2013, Gaithersburg, Maryland, USA, November 19-22, 2013}, volume
  500-302 of \emph{{NIST} Special Publication}. National Institute of Standards
  and Technology {(NIST)}.

\bibitem[{McCoy et~al.(2019)McCoy, Pavlick, and Linzen}]{mccoy-etal-2019-right}
Tom McCoy, Ellie Pavlick, and Tal Linzen. 2019.
\newblock \href {https://doi.org/10.18653/v1/P19-1334} {Right for the wrong
  reasons: Diagnosing syntactic heuristics in natural language inference}.
\newblock In \emph{Proceedings of the 57th Annual Meeting of the Association
  for Computational Linguistics}, pages 3428--3448, Florence, Italy.
  Association for Computational Linguistics.

\bibitem[{Naik et~al.(2018)Naik, Ravichander, Sadeh, Rose, and
  Neubig}]{naik-etal-2018-stress}
Aakanksha Naik, Abhilasha Ravichander, Norman Sadeh, Carolyn Rose, and Graham
  Neubig. 2018.
\newblock \href {https://aclanthology.org/C18-1198} {Stress test evaluation for
  natural language inference}.
\newblock In \emph{Proceedings of the 27th International Conference on
  Computational Linguistics}, pages 2340--2353, Santa Fe, New Mexico, USA.
  Association for Computational Linguistics.

\bibitem[{Niven and Kao(2019)}]{niven_probing_2019}
Timothy Niven and Hung-Yu Kao. 2019.
\newblock Probing {Neural} {Network} {Comprehension} of {Natural} {Language}
  {Arguments}.
\newblock In \emph{Proceedings of the 57th {Annual} {Meeting} of the
  {Association} for {Computational} {Linguistics}}, pages 4658--4664, Florence,
  Italy. Association for Computational Linguistics.

\bibitem[{Pang et~al.(2016)Pang, Lan, Guo, Xu, Wan, and Cheng}]{pang_text_2016}
Liang Pang, Yanyan Lan, Jiafeng Guo, Jun Xu, Shengxian Wan, and Xueqi Cheng.
  2016.
\newblock Text {Matching} as {Image} {Recognition}.
\newblock In \emph{Proceedings of the {Thirtieth} {AAAI} {Conference} on
  {Artificial} {Intelligence}}, {AAAI}'16, pages 2793--2799. AAAI Press.
\newblock Event-place: Phoenix, Arizona.

\bibitem[{Pennington et~al.(2014)Pennington, Socher, and
  Manning}]{pennington_glove_2014}
Jeffrey Pennington, Richard Socher, and Christopher Manning. 2014.
\newblock {GloVe}: {Global} {Vectors} for {Word} {Representation}.
\newblock In \emph{Proceedings of the 2014 {Conference} on {Empirical}
  {Methods} in {Natural} {Language} {Processing} ({EMNLP})}, pages 1532--1543,
  Doha, Qatar. Association for Computational Linguistics.

\bibitem[{Rao et~al.(2016)Rao, He, and Lin}]{10.1145/2983323.2983872}
Jinfeng Rao, Hua He, and Jimmy Lin. 2016.
\newblock Noise-contrastive estimation for answer selection with deep neural
  networks.
\newblock In \emph{Proceedings of the 25th ACM International on Conference on
  Information and Knowledge Management}, CIKM '16, page 1913–1916, New York,
  NY, USA. Association for Computing Machinery.

\bibitem[{Rao et~al.(2019)Rao, Yang, Zhang, Ture, and Lin}]{rao2019multi}
Jinfeng Rao, Wei Yang, Yuhao Zhang, Ferhan Ture, and Jimmy Lin. 2019.
\newblock Multi-perspective relevance matching with hierarchical convnets for
  social media search.
\newblock \emph{Proceedings of the Thirty-Third AAAI Conference on Artificial
  Intelligence (AAAI)}.

\bibitem[{Shen and Lapata(2007)}]{shen-lapata-2007-using}
Dan Shen and Mirella Lapata. 2007.
\newblock Using semantic roles to improve question answering.
\newblock In \emph{Proceedings of the 2007 Joint Conference on Empirical
  Methods in Natural Language Processing and Computational Natural Language
  Learning ({EMNLP}-{C}o{NLL})}, pages 12--21, Prague, Czech Republic.
  Association for Computational Linguistics.

\bibitem[{Vaswani et~al.(2017)Vaswani, Shazeer, Parmar, Uszkoreit, Jones,
  Gomez, Kaiser, and Polosukhin}]{vaswani_attention_2017}
Ashish Vaswani, Noam Shazeer, Niki Parmar, Jakob Uszkoreit, Llion Jones,
  Aidan~N. Gomez, undefinedukasz Kaiser, and Illia Polosukhin. 2017.
\newblock Attention is {All} {You} {Need}.
\newblock In \emph{Proceedings of the 31st {International} {Conference} on
  {Neural} {Information} {Processing} {Systems}}, {NIPS}'17, pages 6000--6010,
  Red Hook, NY, USA. Curran Associates Inc.
\newblock Event-place: Long Beach, California, USA.

\bibitem[{Wan et~al.(2016)Wan, Lan, Guo, Xu, Pang, and
  Cheng}]{10.5555/3016100.3016298}
Shengxian Wan, Yanyan Lan, Jiafeng Guo, Jun Xu, Liang Pang, and Xueqi Cheng.
  2016.
\newblock A deep architecture for semantic matching with multiple positional
  sentence representations.
\newblock In \emph{Proceedings of the Thirtieth AAAI Conference on Artificial
  Intelligence}, AAAI'16, page 2835–2841. AAAI Press.

\bibitem[{Wang et~al.(2018)Wang, Singh, Michael, Hill, Levy, and
  Bowman}]{wang_glue_2018}
Alex Wang, Amanpreet Singh, Julian Michael, Felix Hill, Omer Levy, and Samuel
  Bowman. 2018.
\newblock {GLUE}: {A} {Multi}-{Task} {Benchmark} and {Analysis} {Platform} for
  {Natural} {Language} {Understanding}.
\newblock In \emph{Proceedings of the 2018 {EMNLP} {Workshop} {BlackboxNLP}:
  {Analyzing} and {Interpreting} {Neural} {Networks} for {NLP}}, pages
  353--355, Brussels, Belgium. Association for Computational Linguistics.

\bibitem[{Wang et~al.(2007)Wang, Smith, and Mitamura}]{wang-etal-2007-jeopardy}
Mengqiu Wang, Noah~A. Smith, and Teruko Mitamura. 2007.
\newblock What is the {J}eopardy model? a quasi-synchronous grammar for {QA}.
\newblock In \emph{Proceedings of the 2007 Joint Conference on Empirical
  Methods in Natural Language Processing and Computational Natural Language
  Learning ({EMNLP}-{C}o{NLL})}, pages 22--32, Prague, Czech Republic.
  Association for Computational Linguistics.

\bibitem[{Wang et~al.(2017)Wang, Hamza, and Florian}]{wang_bilateral_2017}
Zhiguo Wang, Wael Hamza, and Radu Florian. 2017.
\newblock Bilateral {Multi}-{Perspective} {Matching} for {Natural} {Language}
  {Sentences}.
\newblock In \emph{Proceedings of the {Twenty}-{Sixth} {International} {Joint}
  {Conference} on {Artificial} {Intelligence}}, pages 4144--4150, Melbourne,
  Australia. International Joint Conferences on Artificial Intelligence
  Organization.

\bibitem[{{Wang} and {Ittycheriah}(2015)}]{2015arXiv150702628W}
Zhiguo {Wang} and Abraham {Ittycheriah}. 2015.
\newblock \href {http://arxiv.org/abs/1507.02628} {{FAQ-based Question
  Answering via Word Alignment}}.
\newblock \emph{arXiv e-prints}, page arXiv:1507.02628.

\end{thebibliography}
\bibliographystyle{acl_natbib}

\appendix
\section{Appendix}
\label{sec:appendix}

\subsection{Datasets Details}\label{app-1}

Here, we provide details for the datasets we use. 

\textbf{Quora Question Pairs (QQP)} \citep{wang_glue_2018} is a widely used benchmark in semantic matching. 
Each pair is annotated with a binary label indicating whether the two texts are paraphrases or not. 
We use split QQP from GLUE \citep{wang_glue_2018} benchmark with \num{363849} examples for training and \num{40430} for testing. 
We report accuracy (Acc) and balanced accuracy (BA).

\textbf{Twitter-URL} \citep{lan_continuously_2017} is a sentence-level paraphrases dataset collected from tweets with \num{42200} examples for training and \num{9334} for testing. 
For each pair, there are \num{6} raw annotations given by human raters. 
We perform the data preprocessing followed the author's notes.$\footnote{https://github.com/lanwuwei/Twitter-URL-Corpus}$
We report macro-F1 and micro-F1.

\textbf{TrecQA} \cite{wang-etal-2007-jeopardy} is a widely used benchmark for question answering. 
According to \citet{10.1145/2983323.2983872}, there are two versions of TrecQA: both have the same training set, but their test sets are different. 
We use the clean version \cite{2015arXiv150702628W} with \num{53417} examples for training and \num{1117} for testing. 
We report mean average precision (MAP) and mean reciprocal rank (MRR).

\textbf{TREC Microblog 2013 (Microblog)}\cite{DBLP:conf/trec/LinE13} is a task to rank candidate tweets by relevance to a short query. 
We use the version prepared by \citet{rao2019multi} with \num{39378} examples for training and \num{6814} for testing. 
We report MAP.

Despite the fact that these datasets differ in tasks (similarity scoring vs. paraphrase detection vs. answer selection vs. tweet search), we regard all of them as a binary classification task to predict the textual similarity between two texts.

\subsection{Models Details}\label{app-2}

Here, we provide details for the models we use. 

\textbf{MatchPyramid} \cite{pang_text_2016} views the matching matrix between two texts as an image, and a CNN is employed to learn hierarchical matching patterns.

\textbf{BiMPM} \cite{wang_bilateral_2017} matches encoded text pairs in two directions with four matching strategies. 
To accelerate the training procedure, we discard the character-composed embedding of the original BiMPM.

\textbf{ESIM} \cite{chen_enhanced_2017} is a sequential inference model based on chain LSTMs. 
We use the base ESIM without ensembling with a TreeLSTM.

\textbf{BERT} \cite{devlin_bert_2019} is a transformer-based \cite{vaswani_attention_2017} pre-trained language model.
Due to the limitation of computational resources, we use BERT$\rm_{\textrm{TINY}}$, which is a compact BERT model with 2 layers and 128 hidden units.

MatchPyramid is based on CNN, BiMPM and ESIM on RNN, and BERT$\rm_{\textrm{TINY}}$ on Transformers.

\subsection{Training Details}\label{app-3}

We provide further details about the evaluation in Section \ref{sec:exp} to facilitate reproducibility.
We implement baselines based on open-source reproduction.$\footnote{The open-source codes are available at https://github.com/pengshuang/Text-Similarity and https://github.com/airkid/MatchPyramid\_torch}$

\textbf{Parameters setting.} For MatchPyramid, BiMPM and ESIM, we use $300$-dimension GloVe word embeddings \cite{pennington_glove_2014}, and keep the pre-trained embeddings fixed during training. 
Words not present in the set of pre-trained words are initialized randomly. 
The kernel size of MatchPyramid is set to be $5\times 5$ and $3\times 3$. 
The dimension of hidden states of ESIM and BiMPM is set to be $128$ and $100$, respectively. 
They are trained using Adam \cite{DBLP:journals/corr/KingmaB14} with initial learning rate of $1e^{-4}$ and batch size of $64$. 
For BERT, we use the implementation provided by the authors$\footnote{Model is available at https://github.com/google-research/bert}$ and apply their default fine-tuning configuration. 

\textbf{Number of parameters in each model.} The number of parameters in MatchPyramid is \num{15053562}, 
of which \num{52962} are trainable. The number of parameters in BiMPM is \num{15890202}, of which \num{889602} are trainable. The number of parameters in ESIM is \num{16695410}, of which \num{1694810} are trainable. The number of parameters in BERT$\rm_{\textrm{TINY}}$ is \num{4386178}, and all of them are trainable.

\end{document}